\documentclass[11pt]{article}

\usepackage[final]{acl}

\usepackage{times}
\usepackage{soul}
\usepackage{latexsym}
\usepackage{multirow}
\usepackage{subcaption}
\usepackage{comment}

\usepackage[T1]{fontenc}

\usepackage[utf8]{inputenc}

\usepackage{microtype}

\usepackage{inconsolata}

\usepackage{graphicx}
\newcommand{\Alona}[1]{\textcolor{blue}{#1}}

\title{Interpretable Humans, Alien LLMs: Expert Analysis of Latent Structures in Assessment Responses\thanks{Accepted for publication at AIME 2026.}}

\author{
Alona Strugatski$^\ast$, 
\bf{Licol Zeinfeld$^\ast$}, 
\bf{Jason Cooper},\\
\bf{Shelley Rap},  
\bf{Gil Schwarts},  
\bf{Giora Alexandron}\\
 Weizmann Institute of Science\\
{\small $^\ast$Equal contribution}\\[2pt]
\{alona.faktor, licol.zeinfeld, jason.cooper, shelley.rap,
gil.schwarts, giora.alexandron@weizmann.ac.il\}\\
}

\begin{document}
\maketitle
\begin{abstract}
The evaluation of large language models (LLMs) relies heavily on human-designed assessments, implicitly assuming that AI and humans employ similar underlying cognitive constructs. Challenging this assumption, we investigate whether the latent factors governing LLM performance carry the same substantive, human-interpretable meaning as the cognitive constructs governing human learners.
Using responses from humans and six LLMs across quantitative reasoning and chemistry assessments, we conducted Exploratory Factor Analysis (EFA) separately for both groups. Subject-Matter Experts (SMEs) then blindly evaluated the resulting factor graphs to ascribe pedagogical meaning to the emerged constructs.
SMEs successfully interpreted \textit{most} of the human-derived factors. Conversely, they could not ascribe meaning to any LLM-derived factors in quantitative reasoning and interpreted only half of the LLM factors in chemistry.
By combining data-driven EFA with blind expert interpretation, this framework shows that LLMs frequently operate on statistically opaque mechanisms distinct from human reasoning.

\end{abstract}

\section{Introduction}

\begin{figure*}[t]
   \centering
   \includegraphics[width=1\textwidth]{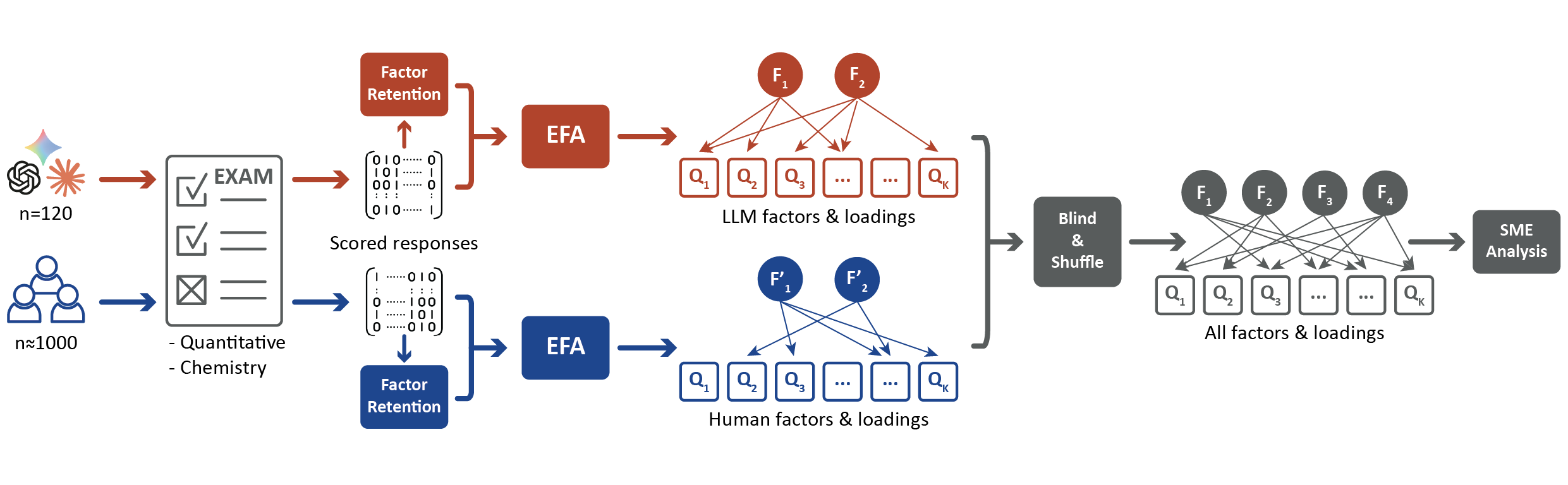}
   \caption{\textbf{Methodological Flowchart. Human and LLM response data for two assessments are processed to extract the underlying factor structure. The blinded and shuffled factor structure results are passed on to subject matter experts for analysis. }}
   \label{fig:flowchart}
\end{figure*}
As large language models (LLMs) are increasingly applied to a wide range of cognitive tasks, the systematic evaluation of their capabilities and limitations has become increasingly important \citep{raiaan2024review, laskar2024systematic}. A common approach to LLM evaluation is using established assessment instruments, particularly standardized exams that were developed for measuring human learners \citep{achiam2023gpt, zhong-etal-2024-agieval}. 
Moreover, such evaluations typically follow the rationale that the assessment that was used to evaluate human skills in a certain domain can be used to make arguments about LLM skills within that domain \citep{katz2024gpt, sanchez-salido-etal-2025-bilingual,jimenez2023swe,borges2024could,Yacobson2026, chlapanis-etal-2025-greekbarbench}. However, this seems LLMs' to make an implicit assumption that LLMs performance on those tests can be attributed to the same skills that humans apply to solve the test items, and that the items-to-skill mapping is similar for humans and LLMs. 

Recent works start to challenge these LLM evaluation practices from educational measurement perspectives, showing, for example, that the structure of factors differs significantly for LLMs and human learners \citep{strugatski2026assessment}. This connects to content-driven analyses that reported that LLM and humans are impacted differently from task dimensions \citep{zeinfeld2026identifying}, and IRT-driven analyses showing that LLM response patterns on educational assessments deviate from the norm and appear as aberrant, compared to human learners \citep{strugatski2025irt,sorenson2024identifying}. This raises questions on what are the factors that govern LLM performance. 

For human respondents, factors are often interpreted in relation to established theories, domain knowledge, and expected response processes \citep{Flora2017DecisionsforVal, leighton2007cognitive}. For LLMs, however, this issue remains under-examined in the literature and is of interest both for understanding LLM capabilities and for designing assessments for contexts in which respondents may use AI (legitimately or not).  Recent work has analyzed LLM response datasets to identify factors that emerge from their response patterns \citep{maimon2025iqtestllmsevaluation}. However, the identified factors were not directly compared with the factors that explain human performance on the same items.  

The present paper addresses this gap by analyzing the latent factors that explain human learners' and LLM performance on the same instruments. It is guided by the following research question (RQ): \\
\noindent\textbf{{RQ}:} How do subject-matter experts interpret the factors derived from human and LLM responses to assessment instruments?

To address this question, we used two established assessment instruments in chemistry and quantitative reasoning, with between a few hundred and a few thousands human responders to each. We then `administered' these instruments to six LLM versions, and computed the factor structure for each respondent group (humans and LLMs) using Exploratory factor analysis (EFA). The result of the EFA was a loading graph for the factors computed for humans and LLMs. We then requested Subject-matter experts (SMEs) to interpret the factors identified for each group and attach pedagogical meaning to them. This was conducted blindly, meaning the SMEs did not know if they are interpreting a factor identified for humans or for LLMs. The research flow is illustrated in Figure 1, and the methodology is detailed in Section~\ref{sec:Method}. 

Previewing the findings, we found that SMEs were able to ascribe meaning to \textit{all} factors revealed for human learners in both quantitative reasoning and chemistry. In contrast, for LLMs, SMEs were unable to ascribe meaning to \textit{any} of the factors revealed in quantitative reasoning, while in chemistry they were able to meaningfully describe only \textit{two of the four} factors. This raises fundamental questions regarding what explains LLM performance, whether their factors are statistical artifacts or simply ones that cannot be explained by humans, and more. 

The key contribution of this study is providing a measurement-oriented perspective on LLM evaluation by moving beyond accuracy-based comparisons to examine the substantive meaning of latent structures derived from assessment responses, and showing that, for the instruments examined, these latent structures not only differed from those of humans but were also largely uninterpretable to human experts.

\section{Related Work}
\subsection{Evaluating LLMs with Human Assessment Instruments}

A common practice in LLM development and evaluation is to reuse instruments originally designed to measure human knowledge or reasoning across domains (e.g., \citep{Kasagga2025MedicalLicensingMeta,Balunovic2025MathArena,zhong-etal-2024-agieval,Du2025SuperGPQA}). This practice is partly driven by the wide availability of such assessments, many of which use closed-form items that are well suited for automatic evaluation. The use of established assessments also provides a common basis for comparing performance across LLMs and, in some cases, against humans \citep{Kasagga2025MedicalLicensingMeta,Yacobson2026}. These comparisons are often treated as evidence of model capabilities in the corresponding domain and are used to expose current capabilities of LLMs. Human-designed assessments are now commonly used in domains such as medical licensing, mathematics, science, legal reasoning, programming, and general academic reasoning, and have been extended across various languages \citep{Balunovic2025MathArena,Zaki2024MaScQA,Arora2023HaveLLMsAdvancedEnough,Chen2025ScholarChemQA,Guo2023ChemLLMBench,Alampara2024MaCBench,zhong-etal-2024-agieval,Du2025SuperGPQA}. 


Despite their practical value, benchmarking LLMs with assessments designed for human learners has several limitations. First, such evaluation may be affected by \textit{contamination}, as publicly available assessments may be included in the LLMs' training data \citep{kapoor2023leakage,xu-etal-2025-contamination,sainz-etal-2023-nlp}. Contamination may result in models performance being the result of memorization rather than reflecting true ability in the domain. 
Second, the evaluations typically report aggregate accuracy scores, which often mask more detailed context-specific variations in performance and undermine the prediction of performance in a real-world setting.
This issue is exacerbated by the limited availability of item-level evaluation results, which are necessary for detailed independent scrutiny of model performance \citep{burnell2023rethinking}.
More fundamentally, equivalent human and LLM assessment performance \textit{does not necessarily imply} that identical underlying constructs are measured, raising a validity concern from educational measurement perspective \citep{mitchell2026evaluating, pmlr-v267-wallach25a}.

Taken together, these limitations suggest that evaluating LLMs with human-designed assessments requires moving beyond aggregate performance comparisons towards finer analyses of response patterns across humans and LLMs.

\subsection{Divergence of Human Learners and LLMs on Educational Assessment}
Increasing evidence indicates that LLMs diverge in multiple aspects from human examinees on educational assessments \citep{Yacobson2026}, that their response patterns appear as aberrant compared to human learners \citep{sorenson2024identifying, strugatski2025irt}, and that their response distributions remain narrow, consistently failing to yield psychometrically plausible, human-like trait profiles even after targeted calibration \citep{petrov2024limited, sauberli-etal-2025-llms, liu2025leveraging}. Additionally, evidence regarding task dimensions shows that specific item features, such as multimodal inputs or under-specified real-world assumptions in STEM contexts, systematically disadvantage models and induce differential item functioning (DIF) between the two populations \citep{borges2024could, wang2024examining, watts2023comparing, zeinfeld2026identifying}. 

\subsection{Educational Measurement Perspectives on LLM Evaluation}
\label{sec:validity}
The core of assessment is construct validity, which  is about establishing a connection between performance on a test and inference on a latent skill it is meant to measure \citep{mislevy2003focus,messick1994interplay}. In educational assessment, the assumption is that the cognitive modeling is known, since developers carefully structure tasks so that a human's performance provides direct evidence of their knowledge. However, evaluating LLMs using assessments designed for humans undermines this basic assumption, since they compute answers using mechanisms that are entirely different from human cognition. Thus, there is a growing need to design evaluation frameworks specifically built for artificial agents \citep{liu-etal-2024-ecbd, maimon2025iqtestllmsevaluation, balepur2025bestdescribesmultiplechoice}. This has led to increasing interest in whether principles from educational and psychological measurement can provide a stronger basis for LLM evaluation \citep{salaudeen2025validityframework,mitchell2026evaluating}.

One possible approach based on educational measurement is to use a data-driven latent structure discovery technique, such as exploratory multidimensional IRT \citep{reckase2009multidimensional}, latent class analysis \citep{collins2010latent} that is useful to identify discrete cognitive processing profiles, and network-based approaches \citep{golino2017exploratory}, which are being adapted to uncover hidden response patterns \citep{munker-2025-fingerprinting}. An additional very common approach is EFA \citep{vandenberg2000review,meredith1993measurement}. EFA is based on estimating factors from patterns of item covariation. In human testing, these statistical tools analyze how responses group together to reveal the underlying cognitive skills driving test performance. 
EFA is becoming increasingly useful in AI evaluation \citep{maimon2025iqtestllmsevaluation,strugatski2026assessment}. However, it does not explain \textit{what} these factors are, and this poses a significant interpretive challenge \citep{Flora2017DecisionsforVal}. 
Experts can interpret these constructs by analyzing item content while also drawing on their understanding of the domain, curricular structures, and learning progressions, and by considering principles of cognitive diagnostic assessment \citep{leighton2007cognitive}.

\section{Methodology}
\label{sec:Method}
\begin{figure*}[t]
   \centering
   \includegraphics[width=1\textwidth]{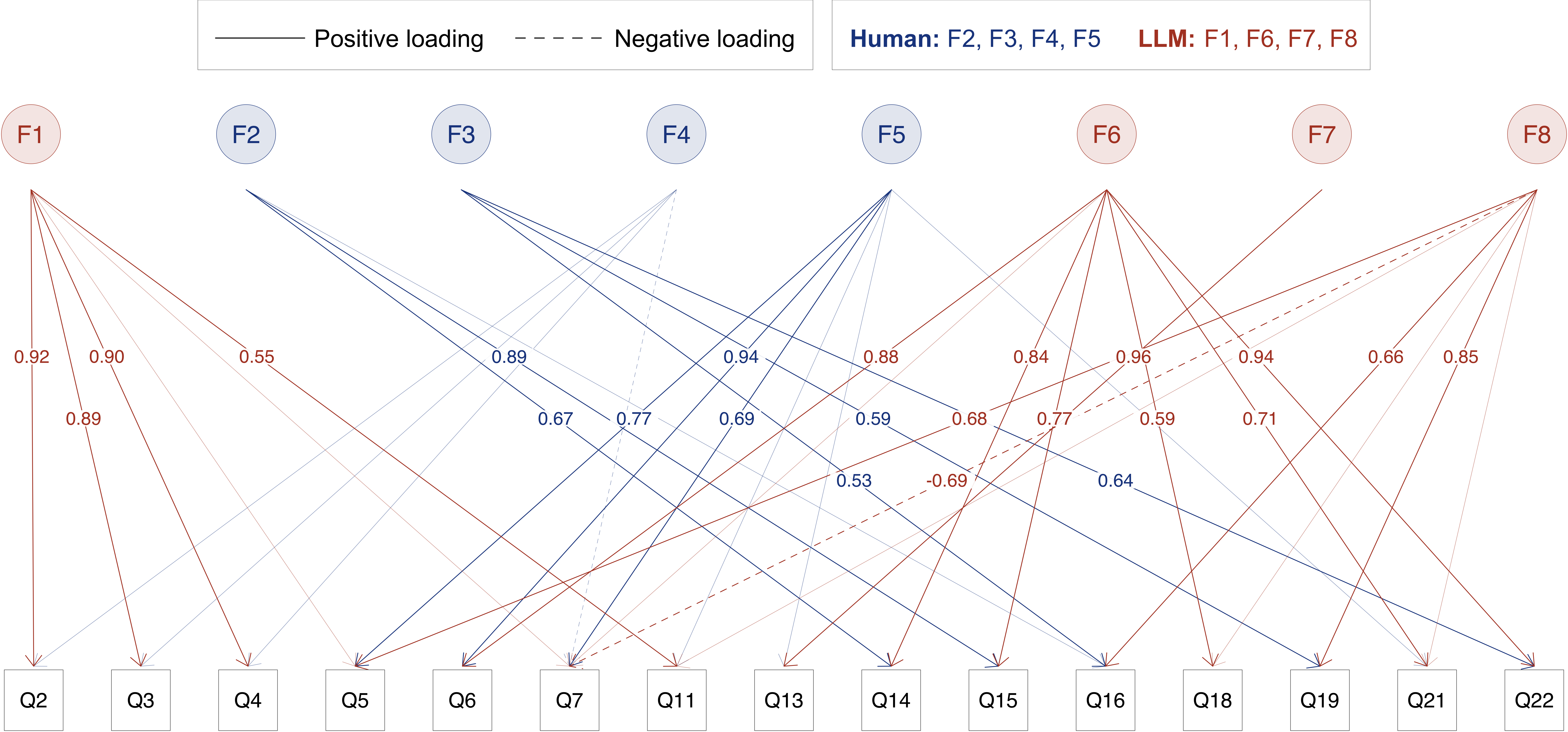}
   \caption{\textbf{EFA item-loading patterns for the chemistry assessment.} Human and LLM factors and loadings are shown in blue and red for visual clarity; however, factor structures were originally extracted independently for human and LLM respondents before being combined, shuffled and blinded into a monochromatic version of this figure in preparation for SME evaluation. $|\mathrm{loadings}| \geq 0.5$ are labeled; $0.3 \leq |\mathrm{loadings}| < 0.5$ are unlabeled; $|\mathrm{loadings}| < 0.3$ not shown.}
   \label{fig:chem_EFA}
\end{figure*}

\begin{figure*}[t]
   \centering
   \includegraphics[width=1\textwidth]{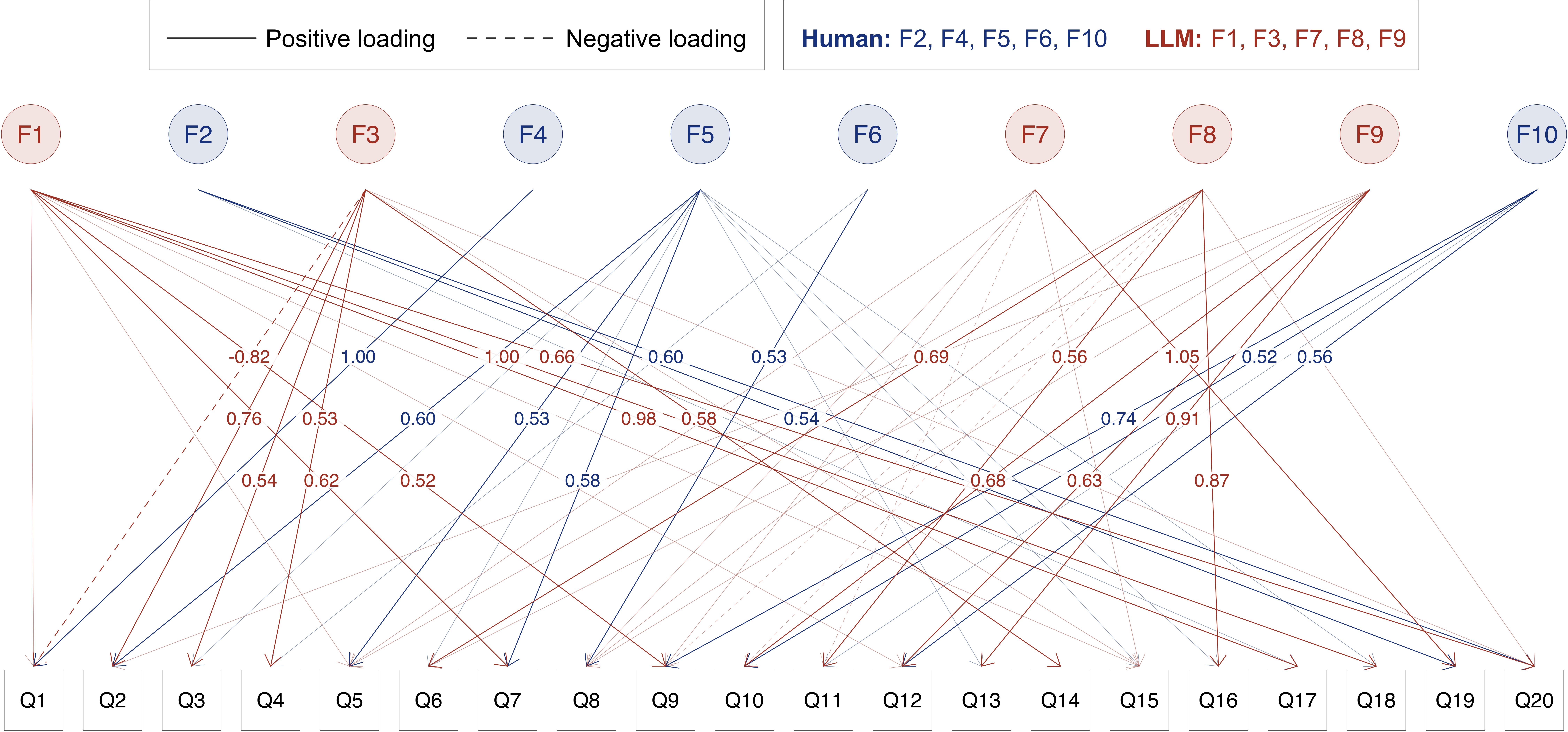}
   \caption{\textbf{EFA item-loading patterns for the quantitative reasoning assessment.} Factor structures independently derived for each group. Visualization and blinding procedures for SME evaluation follow the identical monochromatic protocol utilized for chemistry assessment.}
   \label{fig:quant_EFA}
\end{figure*}

\subsection{Overview}
Our methodology combines exploratory factor analysis (EFA) with blind subject-matter expert (SME) interpretation. The goal of our procedure was to identify latent factors underlying human and LLM response patterns and examine how SMEs characterize those factors based on the items that loaded on them.
As shown in Figure~\ref{fig:flowchart}, the analysis pipeline consisted of six steps. (1) \textbf{Data collection:} Human and LLM item responses were collected for two instruments, yielding separate response datasets for each group. (2) \textbf{Data preparation:} Responses in both datasets were scored as correct (1) or incorrect (0), converted into binary matrices, and preprocessed. (3) \textbf{Applying factor retention criteria:} The number of factors for each response group was determined based on the preprocessed matrices. (4) \textbf{Fitting EFA:} EFA models were fitted separately for humans and LLMs, producing a factor-loading structure for each group based on the factor-retention result. (5) \textbf{Blind \& shuffle:} The EFA results from both groups were grouped, blinded and shuffled in preparation for SME analysis. (6) \textbf{SME interpretation:} SMEs examined the set of items that loaded on to each factor. The full code, sample data, and prompts to reproduce the analyses in the paper can be found here: (\href{https://github.com/LicolZeinfeld/latent_factor_structure_analysis/}{\textcolor{blue}{\underline{link}}}).


\subsection{Data}
\label{sec: Data}
\noindent\textbf{Instruments \& Human Data} In this study, we analyzed responses from two human-designed assessment instruments. (1) A 22-item high-school \textit{chemistry} diagnostic assessment administered through Moodle to students preparing for the matriculation examination. The dataset included responses from 931 students, with an average score of (M=71.49, SD=16.95, out of 100). (2) A 20-item \textit{quantitative reasoning} section from a national university entrance exam, with access to more than 4,800 examinee responses (M=12.45, SD=3.75, out of 20); for the analysis 979 were used. Both assessments included multiple-choice items and contained a mixture of textual and multimodal content, including figures, images, and formulas.

While the data is not publicly available, it provides an advantage of combining high-quality human responses with domain-specific assessments designed and administered in real educational settings.
This combination is essential for SMEs, who depend on evaluating real-world tasks with disciplinary content, and representations.

\noindent\textbf{Collection \& Pooling of LLM Responses} LLM response data were generated using six multimodal models from three major proprietary families: Anthropic’s Claude 3.5 Sonnet and Claude 4.5, OpenAI’s GPT-4o and GPT-5.2, and Google’s Gemini 1.5 Pro and Gemini 3 Pro. Since both assessments contained visual and symbolically rich material, including figures, formulas and disciplinary representations, we restricted data collection to models capable of processing multimodal input. For each model--instrument combination, we obtained 20 independent response sets, resulting in 120 LLM response sets per instrument. Across these pooled responses, LLM performance was (M=76.14, SD=15.41) on the chemistry instrument and (M=11.94, SD=4.52) on the quantitative reasoning instrument.

The generated responses were then pooled across the six models into a single LLM respondent group, consistent with recent NLP evaluation work that analyzes performance across multiple models \citep{macko-etal-2023-multitude}. This decision follows from the goal of the study to compare the latent response structures that emerge for human respondents and for LLMs broadly, rather than to characterize model-specific factor structures. Additionally, pooling increased variation in LLM response patterns, consistent with the human data variation, allowing for stable EFA fit results. Conceptually, pooling introduces heterogeneity into the data, aligning with EFA’s assumption that human respondents exhibit heterogeneous response patterns.


\noindent\textbf{LLM Prompting Technique}
To prevent context leakage, we collected responses via the models' web interfaces using independent, temporary chat sessions. Mirroring authentic human test administration, we provided the complete instrument as a single PDF rather than item-by-item, instructing the models to output only their final selected answers. 
Following \citep{munker-2025-fingerprinting}, we adopted a minimal, zero-shot prompting strategy to elicit direct LLM responses. Since LLM outputs can vary substantially with prompt engineering \citep{zhuo-etal-2024-prosa}, we intentionally avoided prompt engineering that would, in our case, introduce an additional source of difficult-to-control variation. 

\subsection{Exploratory Factor Analysis}
\label{subsec:EFA}
\noindent\textbf{Data preparation}
\label{subsec:preproc} 

\textbf{(1) Response Scoring}
Finally, model responses were scored against the answer key and converted to binary outputs. As with the human response data, omitted, malformed, or otherwise invalid answers were treated as incorrect.
\textbf{(2) Preprocessing}
Since factor-retention and EFA procedures require continuous distributions for item-responses, we used the binary item-response data to estimate these distributions. As a preprocessing step, we screened the item sets and removed items that could produce unstable estimates. We first excluded any items with zero variance in either human or LLM group. Then, we inspected the \(2 \times 2\) contingency tables underlying each item-pairwise correlation and removed items contributing to empty or highly sparse cells. As a result, no items were excluded from the quantitative reasoning item set. Seven items were excluded from the chemistry item set: 1, 8--10, 12, 17, and 20. All subsequent analyses were conducted on the common retained item set for humans and LLMs.

\noindent\textbf{Factor-retention}
We specified the dimensionality of the EFA models by applying the Kaiser criterion (eigenvalues > 1) \citep{Sunbok2017FA, Nazaretsky2022Trust, Kaiser1974} independently to the human and LLM tetrachoric correlation matrices. This retention method yielded an identical number of factors for both populations within each instrument: four for chemistry and five for quantitative reasoning, establishing the structural baseline needed to examine their respective item-loading patterns.

\noindent\textbf{Factor-extraction}
Using the retained number of factors, we fit EFA models separately for the human and LLM response groups. Factor extraction was conducted with minimum residual estimation in the \texttt{psych} package, using \texttt{fa} with \texttt{fm = "minres"}, which seeks a solution that closely reproduces the observed correlation matrix. Because the extracted factors were not assumed to be independent, we applied an oblique rotation using \texttt{rotate = 
"oblimin"}~\citep{revelle2025psych}.
Figures~\ref{fig:chem_EFA} and ~\ref{fig:quant_EFA} present the EFA loading structure results for the human and LLM response groups for both instruments using the Kaiser-retained factor solutions and table \ref{tab:efa_results} the quality of fit matrices for human (all \texttt{RMSEA} < 0.02 ) and LLM (all \texttt{RMSEA} < 0.1 ) with confidence (> 0.90) for both, representing a good model fit for humans and marginal for LLMs.

While an identical number of factors were used for both groups (table\ref{tab:efa_results}), the underlying item-loading structures diverged \citep{strugatski2026assessment}. For example, items Q5, Q16, and Q19 loaded heavily (> 0.5) onto a single LLM factor (F8), yet split across two distinct human factors (F2 and F5). This structural misalignment also permeated quantitative reasoning assessments (see \ref{fig:quant_EFA}). Next, we examine how SMEs interpreted these factor structures.

\begin{table*}[t]
\centering
\caption{Factor Retention and EFA Fit for Human and LLM Samples}
\label{tab:efa_results}
\begin{tabular}{|l|cc|cc|cc|}
\hline
& \multicolumn{2}{c|}{\textbf{Factors Retained}} & \multicolumn{4}{c|}{\textbf{EFA Fit Results}} \\
\cline{2-7}
& & & \multicolumn{2}{c|}{\textbf{Human}} & \multicolumn{2}{c|}{\textbf{LLM}} \\
\cline{4-7}
\textbf{Instrument} & \textbf{Human} & \textbf{LLM} & \textbf{RMSEA} & \textbf{Confidence} & \textbf{RMSEA} & \textbf{Confidence} \\
\hline
Chemistry              & 4 & 4 & 0.019 & $> 0.90$ & 0.045 & $> 0.90$ \\
Quantitative reasoning & 5 & 5 & 0.011 & $> 0.90$      & 0.092 & $> 0.90$ \\
\hline
\end{tabular}
\end{table*}

\subsection{SME Analysis} 
\label{subsec:factor-similarity-analysis}
Experts in each of the fields -- mathematics and chemistry -- were asked to try to explain the factors that emerged from EFA~\citep{Sireci1998ContentValidity}. To avoid bias, the factors were blinded in the sense that they were randomly assigned numbers from 1 to 10, obscuring which reflect human responses and which reflect LLM responses. 
The instruction was: What do the items that loaded highly on each factor have in common, and in what ways are they different from other items? The experts were instructed to consider three kinds of characteristics of the items: epistemic (e.g. curricular content), cognitive (e.g. order of thinking required to solve) and psychometric (e.g. effectiveness of elimination strategies). Experts ordered the items in each factor and asked themselves what the first two have in common, what the third has in common with them, and so on. If and when common characteristics were found, the items outside the factor were analyzed with respect to these characteristics. The quality of an explaining characteristic was measured in terms of the number of items that load strongly on the factor and hold the characteristic and the number of items outside the factor that do not.

\section{SME Analysis Results}
\subsection{Quantitative Reasoning Factors}
In the quantitative reasoning instrument, whose factor structure is shown in Figure \ref{fig:quant_EFA}, none of the LLM factors could be explained by the experts in a satisfactory manner. Three human factors were explained, as follows:
Factor F10 (epistemic): Reading data from a graph. Items Q9-12 all refer to data presented graphically, and they all loaded on this factor, and only on it (Q9, Q10 and Q12 strongly, Q11 weakly).
Factor F5 (psychometric): It is possible to check the correctness of each offered answer without solving the problem constructively. This is a strategy that is often advocated in preparing for the psychometric test. In each of the items that loaded strongly on this factor (Q2, Q5, Q7) it is helpful to substitute the offered answers for the unknown and to check for each if it is a correct answer. This is also the case for some of the items that loaded weakly. For example, in item Q3, substituting the smallest answer (0) and recognizing that this is a possible value for $(x-y)^2$ solves the problem. 
Factor F2 (cognitive/psychometric): It is possible to solve the general problem by working with an imagined example. This is a well known psychometrics strategy, which is sometimes called autonomous concrete instantiation (as opposed to substitution of the given answers) \citep{Koedinger2004Story}. In item Q20, this involves arbitrarily selecting three prime numbers and working with their product and, in item Q19, imagining the vertices of an arbitrary equilateral triangle and trying to imagine a fourth point for which the condition holds. Item Q17, which loaded weakly on this factor, does not seem to  have this characteristic. Item Q13, which did not load on this factor, may seem to have this characteristic (work with arbitrary A, B, and C for which the given holds); however, there the question is which claim holds necessarily, hence an arbitrarily selected example is not generic and will generally not lead to the correct answer.

\subsection{Chemistry Factors}
To better understand the chemistry instrument's underlying factor structure, shown in Figure \ref{fig:chem_EFA}, an initial attempt was made to identify commonalities based on chemistry-related content. However, the factor structure could not be adequately explained on the basis of content alone. Instead, several similarities were identified in the structural characteristics of the items rather than in their substantive content. One exception to this pattern is discussed separately below, as it remains unclear whether the factor emerged due to the nature of the items themselves or due to their shared content. Factors F1 and F4, which primarily included Items Q2, Q3, and Q4, appeared to group together because these items consisted of subsections in which all correct responses were selected from only two possible answer options. This shared response structure may account for the observed factor loading. Nevertheless, it is important to note that additional multiple-choice and sentence-completion items did not load onto this factor, suggesting that response format alone does not fully explain the pattern. Factor F2 consisted of two items that were sequentially related, with one item serving as a continuation of the other. This close structural and contextual relationship likely contributed to their loading onto the same factor. Factors F3 and F8 included items Q16 and Q19, both of which incorporated a true/false response format. The shared evaluative structure of these items may explain their association within the factor analysis. Factor F5 may be interpreted, at least partially, on the basis of content. The items associated with this factor addressed dissolution processes at the microscopic level. At the same time, items Q6 and Q7 also demonstrated substantial visual and procedural similarity, as they required a highly comparable mode of response. In contrast, no clear explanation could be identified for factor F6. Factor F7 was not ascribed a meaning since only one item loaded on it.

\section{Discussion}
The results show that, on the instruments examined, not only that LLMs use different `skills' to answer items, but that these skill may not be human interpreted. The fact that in most cases the SMEs did not find any logic in the LLMs' reasoning indicates a misalignment between how SMEs and LLMs organize domain knowledge. And if LLM factors are largely opaque for SMEs, it becomes unclear what are the constructs that we are measuring for LLMs anyway. This provides a different explanation to why LLMs sometimes perform well on a benchmark but poor on tasks that SMEs see as requiring the same skill set \citep{Kasagga2025MedicalLicensingMeta, ullman2023large}, and extends recent studies arguing that making inference about LLM skills based on assessments designed for evaluating these skills among human learners may be ungrounded \citep{pmlr-v267-wallach25a, inger2025forget, ullman2023large}.


Another possibility explaining SMEs' failure to find meaning in the LLM factors is that their factors represent non-human skills. But a different possibility is that the LLMs behavior is merely the result of a statistical process that cannot be meaningfully explained as resulting from latent traits. Delving into this question may require deeper investigation into how LLMs represent domain knowledge and activate it during problem solving. 

This leads to a broader question. A widely held view in neuroscience is that the basic organization of the brain is broadly similar across individuals \citep{kanwisher2010functional,yeo2011organization}, and that differences in abilities arise in large part from environmental factors, such as the learning opportunities individuals encounter throughout their lives \citep{noble2015family}. In contrast, LLMs differ substantially in their underlying architectures, in addition to ``environmental'' differences related to their ``learning opportunities''. Given this variability in underlying mechanisms, an interesting question is whether a unified definition and measurement of abilities -- independent of the mechanisms that underlie them -- is even feasible.

\noindent{\textbf{\textit{Limitations.}}}
We acknowledge several limitations in our data analysis that should be considered when interpreting our findings. First, using non-public datasets constrains reproducibility, but was important for reducing the risk of contamination, which could have bias the results. In terms of internal validity, the LLMs dataset is relatively small compared to the human sample. We also pooled different LLMs together, assuming that they can be treated as a single population, despite differences in their underlying architectures. This is an accepted, yet debatable practice \citep{maimon2025iqtestllmsevaluation, munker-2025-fingerprinting}. The analysis itself was conducted by a small number of SMEs and did not include an inter-rater agreement process. In terms of external validity, the research is based on a small number of instruments, and was produced in a single language. 

\noindent{\textbf{\textit{Implications to educational applications.}}} If students and LLMs do not share the same latent structure on a task, their use in educational applications that rely on a common interpretation of the meaning space may hinder their usability as personalized tutors, for assessment design, and for simulating students for purposes such as creating learning pals.

\noindent\textbf{\textit{Contribution \& Future Work}.}
This work makes two primary contributions. First, we present a methodological framework to examine the substantive meaning of latent structures derived from assessment responses by combining EFA with blind SME interpretation. Second, we contribute empirical findings highlighting the inherent difficulty human experts face when attempting to ascribe substantive meaning to the latent constructs that emerge from LLMs. In future work, we will upscale the scope of assessment tools to include more modalities and a larger volume of data, including open-source datasets. We also plan to explore the pooling effect by examining the emerged constructs of each model separately.






\section*{Acknowledgments}
This work was supported by the Knell Family Institute for Artificial Intelligence, Israel. The authors thank the National Institute for Testing and Evaluation for providing access to psychometric exam data.


\bibliography{validity}
\end{document}